\documentclass[11pt,letterpaper]{article}
\usepackage{ijcnlp2017}
\usepackage{times}
\usepackage{latexsym}

\ijcnlpfinalcopy

\def\ijcnlppaperid{1105}

\usepackage{cuted, ragged2e}
\usepackage{lipsum, array, booktabs, caption}
\usepackage{dblfloatfix}
\usepackage{latexsym,graphicx}
\usepackage{amsmath}

\title{Learning to Explain Non-Standard English Words and Phrases}

\author{Ke Ni\and William Yang Wang\\
  Department of Computer Science\\
  University of California, Santa Barbara\\   Santa Barbara, CA 93106 USA\\
  {\tt \{ke00@umail\},\{william@cs\}.ucsb.edu}}

\date{}

\begin{document}

\maketitle

\begin{abstract}
We describe a data-driven approach for automatically explaining new, non-standard English expressions in a given sentence, building on a large dataset that includes 15 years of crowdsourced examples from \emph{UrbanDictionary.com}.
Unlike prior studies that focus on matching keywords from a slang dictionary, we investigate the possibility of learning a neural sequence-to-sequence model that generates explanations of unseen non-standard English expressions given context.
We propose a dual encoder approach---a word-level encoder learns the representation of context, and a second character-level encoder to learn the hidden representation of the target non-standard expression.
Our model can produce reasonable definitions of new non-standard English expressions given their context with certain confidence.
\end{abstract}

\section{Introduction}
In the past two decades, the majority of NLP research 
focused on developing tools for the Standard English on newswire data. However, the non-standard part of the language is not well-studied in the community, even though it becomes more and more important in the era of social media.
While we agree that one must take a cautious approach to
automatic generation of non-standard language~\cite{Hickman2013},
but for many practical purposes,
it is also of crucial importance for machines to be able to \emph{understand} and \emph{explain} this important subversion of the language.

In the NLP community, 
using dictionaries of non-standard language as an external knowledge source is useful for many tasks. For example, 
Burfoot and Baldwin~\shortcite{burfoot2009automatic}
consult the slang definitions from Wiktionary to
detect satirical news articles.
Wang and McKeown~\shortcite{wang-mckeown:2010:PAPERS} show that using a slang dictionary of 5K terms can help detecting vandalism of Wikipedia edits. 
Han and Baldwin~\shortcite{han2011lexical} make use of the same slang dictionary, and achieve the best performance by combining the dictionary lookup approach with word similarity and context for Twitter and SMS text normalization.
However, using a 5K slang dictionary may suffer from the coverage issue,  since slang is evolving rapidly in the social media age\footnote{For example,
more than 2K entries are
submitted daily to Urban Dictionary~\cite{Kolt2009},
the largest online slang resource.}.
\begin{figure}[t]
\includegraphics[scale=0.27]{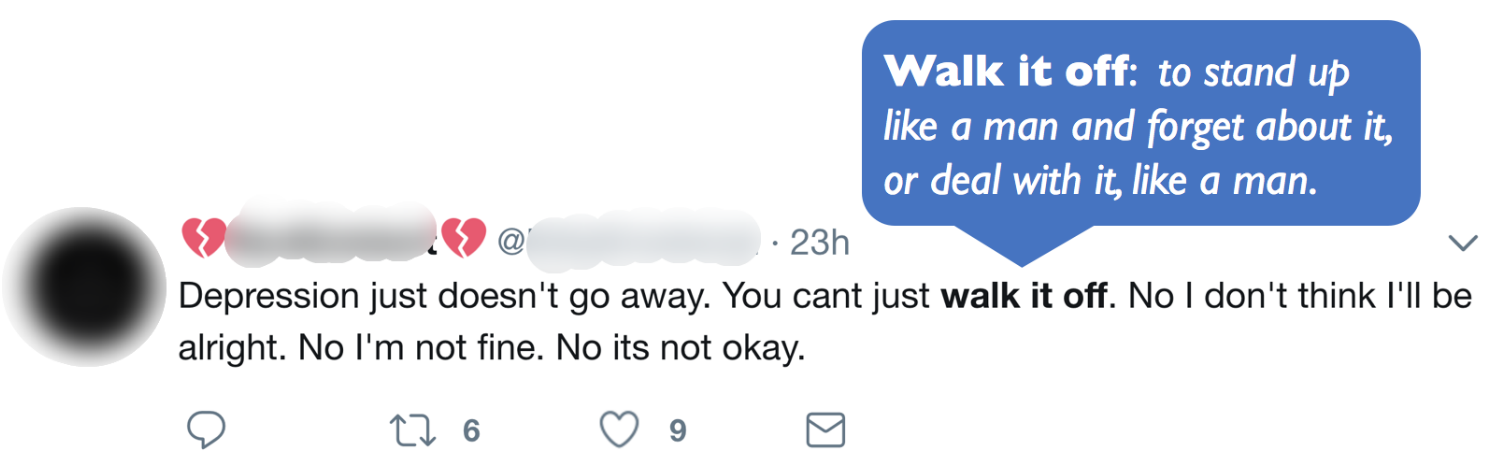}  
\caption{An example Tweet with a non-standard English expression. Our model aims at automatically explaining any newly emerged, non-standard expressions (generating the blue box).}
\label{fig:example}
\vspace{-1ex}
\end{figure}
Recently, Thanapon Noraset~\shortcite{Norasetn:2016:PAPERS} shows that it is possible to use word embeddings to generate plausible definitions. Nonetheless, one weakness is that definition of a word may change within different contexts.

In contrast, we take a more radical approach: we aim at building a general purpose sequence-to-sequence neural network model~\cite{sutskever2014sequence} to explain any non-standard English expression, which can be used in many NLP and social media applications. More specifically, given a sentence that includes a non-standard English expression, we aim at automatically generating the translation of the target expression. Previously, this is not possible because
the resources of labeled non-standard expressions are not available. In this paper, we collect
a large corpus of 15 years of crowdsourced examples,
and formulate the task as a sequence-to-sequence generation problem.
Using a word-level model, we show that it is possible to build a general purpose non-standard English words and phrases explainer using neural sequence learning techniques.  
%In experiments, we show that our approach significantly outperforms competitive baselines, and the improvements are consistent across various settings. 
To summarize our contributions:
\begin{itemize}
\item We present a large, publicly available corpus of non-standard English words and phrases, including 15 years of definitions and examples for each entry via crowdsourcing;
\item We present a hybrid word-character sequence-to-sequence model that directly explains unseen non-standard expressions from social media;
\item Our novel dual encoder LSTM model outperforms a standard attentive LSTM baseline, and it is capable of generative plausible explanation for unseen non-standard words and phrases.
\end{itemize}
In the next section, we outline related work on non-standard expressions and social media text processing. We will then introduce our dual encoder based attentive sequence-to-sequence model for explaining non-standard expressions in Section~\ref{sec:model}. Experimental results are shown in Section~\ref{sec:exp}. And finally, we conclude in Section~\ref{sec:conclude}.

\section{Related Work}
\label{sec:related}
The study of non-standard language is of interests to many researchers in the social media and NLP communities. For example, Eisenstein et al.~\shortcite{eisenstein2010latent} propose a latent variable model to study the lexical variation of the language on Twitter,
where many regional slang words are discovered.
Zappavigna~\shortcite{zappavigna2012discourse}
identifies the Internet slang as an important component in the discourse of Twitter and social media. Gouws et al.~\shortcite{gouws2011contextual}
provide a contextual analysis of how social media users shorten their messages.
Notably, a study on Tweet normalization~\cite{han2011lexical} finds that, even when using a small slang dictionary of 5K words, Slang makes up 12\% of the ill-formed words in a Twitter corpus of 449 posts.
In the NLP community, slang dictionary is widely used in many tasks and applications~\cite{burfoot2009automatic,wang-mckeown:2010:PAPERS,rosenthal2011age}.
However, we argue that using a small, fixed-size dictionary approach may be suboptimal: 
it suffers from the low coverage problem, and to keep the dictionary up to date, maintaining such dictionary is also expensive and time-consuming.
To the best of our knowledge, our work is the first to build a general purpose machine learning model for explaining non-standard English terms, using a large crowdsourced dataset.

\section{Our Approach}
\label{sec:model}
\subsection{Sequence-to-Sequence Model}
\begin{figure}[t]
\centering
\includegraphics[scale=0.3]{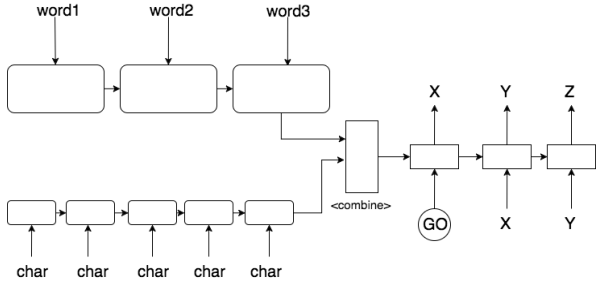}  
\caption{Dual Encoder Structure. \emph{Top left: a word-level LSTM encoder for modeling context. Bottom left: a character-level LSTM encoder for modeling target non-standard expression. Right: an LSTM decoder for generating explanations.}}
\label{fig:model}
\vspace{-1ex}
\end{figure}

Since our goal is to automatically generate explanations for any non-standard English expressions, we select sequence-to-sequence models with attention mechanisms as our fundamental framework~\cite{bahdanau2014neural}, which can produce abstractive explanations, and assign different weights to different parts of a sentence. To model both the context words and the non-standard expression, we propose a hybrid word-character dual encoder. An overview of our model is shown in Figure~\ref{fig:model}.

\subsection{Context Encoder}
Our context encoder is basically a recurrent neural network with long-short term memory (LSTM)~\cite{hochreiter1997long}. LSTM consists of input gates, forget gates and output gates, together with hidden units as its internal memory. Here, $i$ controls the impact of new input, while $f$ is a forget gate, and $o$ is an output gate. $\tilde{C_t}$ is the candidate new value.
$h$ is the hidden state, and $m$ is the cell memory state. ``$\odot$'' means element-wise vector product. The definition of the gates, memory, and hidden state is:
\[i_t = \sigma(W_i [x_t, h_{t-1}])\]
\[f_t = \sigma(W_f [x_t, h_{t-1}])\]
\[o_t = \sigma(W_o [x_t, h_{t-1}])\]
\[\tilde{C_t} = \tanh(W_c [x_t, h_{t-1}]) \]
\[m_t = m_{t-1} \odot f_t + i_t \odot \tilde{C_t}\]
\[h_t = m_t \odot o_t\]

\noindent At each step, RNN is given a vector as input, changes its hidden states and produces outputs from its last layer. Hidden states and outputs are stored and later passed to a decoder, which produces final outputs based on hidden states, outputs, and the decoder's own parameters.The context encoder learns and encodes sentence-level information for the decoder to generate explanations.

\subsection{Attention Mechanism}
The idea of designing an attention mechanism is to select key focuses in the sentence. More specifically, we would like to pay attention to specific parts of encoder outputs and states. We follow a recent study~\cite{vinyals2015grammar} to setup an attention mechanism. We have a separate LSTM for decoding.

Here we briefly explain the method. When the decoder starts its decoding process, it has all hidden units from the encoder, denoted as $(h_1..h_{T_1})$. Also we denote the hidden state of the decoder as $(d_1..d_{T_2})$. $T_1$ and $T_2$ are input and output lengths. At each time step, the model computes new weighted hidden states based on encoder states and three learnable components: $v$, $W'_1$ and $W'_2$. 
\[u^t_i=v^T tanh(W'_1 h_i + W'_2d_t)\]
\[a^t_i=softmax(u^t_i)\]
\[d'_t = \sum_{i=1}^{T_1} a^t_i h_i\]

\noindent Here $a$ denotes the attention weights. $u^t$ has length $T_1$. After the model computes $d'_t$, it concatenates $d'_t$ and $d_t$ as new hidden states used for prediction and next update.

\iffalse
\subsection{Pure Character-level Model}
Because out-of-vocabulary words show up frequently in data, we consider using character-level models to better handle these words. Our approach is to simply tokenize sentences and feed them into the same model. One of the limitations is that we need larger memory space because input lengths and output lengths are much longer than before. Other issues will be presented in the experiment section.
\fi

\subsection{Dual Encoder Structure}
Given a single context encoder, it is challenging for any decoder to generate explanation in the instance. The reason is that there could be multiple non-standard expressions in the sentence, and it confuses the decoder on which one it should explain. In practice, the user can often pin point to the exact expression she or he does not know, so in this work, we design a second encoder to learn the representation of the target non-standard expression.

Since our goal is to explain any non-standard English words and phrases, we consider a character-level encoder for this purpose. This second encoder reads the embedding vector of each character at each time step, and produces an output vector, and hidden states. Our dual encoder model linearly combines the hidden states of the character-level target expression encoder with the hidden states of the word-level encoder with the following equation:
\[ h_{new} = h_1 W_1 + h_2 W_2 + B\]

\noindent Here, $h_1$ is the context representation, and $h_2$ is the target expression representation. The Bias $B$ and the combination weights $W_1$ and $W_2$ are learnable.

\section{Experiment}
\label{sec:exp}
\subsection{Dataset}
We collect the non-standard English corpus\footnote{We have released the dataset for research usage. The dataset is available at:~\url{http://www.cs.ucsb.edu/~william/data/slang_ijcnlp.zip}}
from Urban Dictionary (UD)\footnote{www.urbandictionary.com}---the largest online slang dictionary,
where terms, definitions and examples
are submitted by the crowd.
UD is made a reliable resource,
due to the quality control of its publishing procedure. To prevent vandalism,
a user must have a Facebook or Gmail account, and 
when each user submits an entry, the UD editors will vote ``Publish'' or ``Don't Publish''~\cite{Lloyd2011}. Each editor is also distinguished by IP addresses,
and HTTP cookies are used to prevent each editor from cheating.
In recent years, United States Federal government has consulted
UD for the definition of ``murk'' in a threat case
~\cite{Jackson2011},
whereas UD is also referred
in a financial restitution case in Wisconsin~\cite{Kaufman2013},
as well as determining appropriate license plates 
in Las Vegas~\cite{Davis2011}.

A total of 421K entries (words and phrases) from the period of 1999-2014 are collected. Each entry includes 
a list of candidate definitions and examples, as well as the surface form of the target term.
Using the UD API, we can pinpoint the token positions of the words/phrases, and obtain the ground truth labels for tagging.

Our training and testing data use all the examples in an entry (a non-standard expression). We randomly select 371,028 entries for training, resulting in 907,624 sequence pairs of instance and reference explanation. The test set includes 50,000 entries, and 61,330 sentences. Note that all testing target non-standard expressions, instances and examples do not overlap with those in the training dataset.

\begin{figure}[t!]
\centering
\small
{\fontfamily{put}\selectfont
\fbox{\parbox[t][4in]{2.7in}{\textbf{Instance}: \emph{``dude taht roflcopter was pretty loltastic!!1!''} \\ 
\textbf{Target}: loltastic \\ 
\textbf{Reference}: something funnyishly fantastic. \\
\textbf{Generated Explanation (Single)}: a really cool , amazing , and good looking .\\
\textbf{Generated Explanation (Dual)}: a word that is extremely awesome .  
\vspace{1.5ex}
\hrule
\vspace{1.5ex}
\textbf{Instance}: \emph{``danny is so jelouse of my work!''} \\ 
\textbf{Target}: jelouse   \\ 
\textbf{Reference}: how unintelligent people who think they are better than someone spells "jealous".    \\
\textbf{Generated Explanation (Single)}: your friend ' s way of saying "     \\
\textbf{Generated Explanation (Dual)}: a word used to describe a situation , or a person who is a complete idiot . 
\vspace{1.5ex}
\hrule
\vspace{1.5ex}

\textbf{Instance}: \emph{``that sir right there, is being quite adoucheous''} \\ 
\textbf{Target}: adoucheous  \\ 
\textbf{Reference}: a person acting in a conformative manner that causes social upset and violence. \\
\textbf{Generated Explanation (Single)}: when a male is being a male and a male . \\
\textbf{Generated Explanation (Dual)}: the act of being a douchebag . 
}}}
\caption{Some generated explanations from our system.}
\label{fig:example}
\vspace{-2ex}
\end{figure}

\subsection{Experimental Settings}
Our implementation is based on Tensorflow\footnote{www.tensorflow.org}. For input embeddings, we randomly initialize the word vectors. We use stochastic gradient descent with adaptive learning rate to optimize the cross entropy function. We use BLEU scores~\cite{papineni2002bleu} for the evaluation.

\subsection{Quantitative and Qualitative Results}
Quantitative experimental results are showed in Table~\ref{tab:results}. Here we compare the performance of our proposed large dual encoder model to a single context encoder, word-level attentive sequence-to-sequence, as well as a full character-level context encoder model. We use 256 hidden units for this full character-level, because it is the largest setting that fits our Titan X Pascale GPU. Character-level model has longer sequence, which becomes one of its shortages compared with word-level model.

We see that the single encoder and full character level context models do not offer the best empirical performances on this dataset. Our novel dual encoder method, which combines the strengths of word-level and character-level sequence-to-sequence models, obtained the best performance.

For qualitative analysis, we provide some generated explanations in Figure~\ref{fig:example}. For example, the first target non-standard expression is the word ``loltastic'', which is a combination of the words ``lol'' and ``fantastic''. To explain this composite non-standard expression, a character-level encoder is needed. The generated explanation from our dual encoder approach clearly makes more sense than the single decoder result.

Overall, dual encoder can explain words with more confidence partly because it knows which words it needs to explain, especially for sentences containing multiple non-standard English words and phrases. Our model can also accurately explain many known acronyms. We also notice that LSTM cells outperform gated recurrent units (GRUs)~\cite{cho2014properties}, and attention mechanism improves the performance.

\begin{table}[t]
  \small
  \centering
  \label{tab:title}
  \begin{tabular}{ |l|l|c|c|c| }
    \hline
    Model (w. attention) & hidden units & B1 & B2\\ 
    \hline
    single encoder
    & 1024 & 21.06 & 2.1\\
    small dual encoder
    & 512 & 21.84 & 2.2\\
    large dual encoder
    & 1024 & \textbf{24.58} & \textbf{2.37}\\
    full char-level model
    & 256 & 21.13 & 1.8\\
    %model3 & simplified LSTM & 1024  & 18.25 & 1.5\\
    \hline
  \end{tabular}
  \newline
  \caption{BLEU scores for explaining non-standard English words and phrases in test dataset.}
  \label{tab:results}
\vspace{-2ex}
\end{table}

\iffalse
\subsection{Issues}
We do not show any result of character level model here but its result is in table 1. Unigram score of character level model is high, but there are too many repeated sentences in our character level model's result. We think the reason is that the model tried to "cheat" by repeating high frequent characters to reduce loss. This also happens in word level models but is less severe. In addition, our models learn generic explanation first, such as "a person who is", "the act of" and "a place where". However, models do give different latter half based on contexts of sentences based on our observation.

\subsection{Impacts of Hyperparameters}
We also have test results for slangs showing up in training data and vocabulary but with different examples in test data. All of our models can explain these words much better and have a higher BLEU score(The best model can achieve 6, 2-gram BLEU score, on the whole test dataset). Furthermore, dual encoder model gives more meaningful results according to our observation. Because of length limitation, we cannot show results here. From other experiment results, we briefly provide some information about hyperparameters: LSTM outperforms GRU. Larger neural network(1024 hidden units) outperforms smaller ones(512, 256 hidden units). Attention mechanism enhances much performance. Pretrained word-embeddings improves OpenNMT results.
\fi

\section{Conclusion}
\label{sec:conclude}
%With recent advances in sequence-to-sequence models, we believe that it is possible to consider training neural network models, and explain the new non-standard English expressions in social media. 
In this paper, we introduce a new task of learning to explain newly emerged, non-standard English words and phrases. To do this, we collected 15-year of UrbanDictionary data, and designed a dual encoder attentive sequence-to-sequence model to learn the hidden context representation and the hidden non-standard expression embedding. We showed that combining word-level and character-level models improved the performance for this task

\newpage

\bibliography{all}
\bibliographystyle{ijcnlp2017}

\end{document}